\documentclass[10pt,twocolumn]{article}

\usepackage[margin=0.75in,columnsep=0.25in]{geometry}
\usepackage[T1]{fontenc}
\usepackage{tgtermes}
\usepackage{amsmath,amssymb}
\usepackage{graphicx}
\usepackage{booktabs}
\usepackage{array}
\usepackage{longtable}
\usepackage{multirow}
\usepackage{caption}
\usepackage[numbers,sort&compress]{natbib}
\usepackage{titlesec}
\usepackage{enumitem}
\usepackage{adjustbox}
\usepackage{xcolor}
\usepackage{float}
\usepackage[section]{placeins}
\usepackage[colorlinks=true,linkcolor=blue!55!black,citecolor=blue!55!black,
  urlcolor=blue!65!black,breaklinks=true]{hyperref}
\usepackage{url}
\titlespacing*{\section}{0pt}{10pt}{6pt}
\titlespacing*{\subsection}{0pt}{8pt}{4pt}
\newcommand{\mter}{MTER}

\title{\Large \textbf{Text-Only Domain Adaptation for Low-Resource Medical ASR:\\
Shallow Fusion and Density Ratio Rescoring for Hausa}}

\author{%
  Oche David Ankeli$^{*1}$, Hassan Adelani Luqman$^{1}$, Theodora Ngozi Egbunike$^{1}$,\\
  Joshua Gottschalk$^{2}$, and Bonaventure F.\ P.\ Dossou$^{3}$\\[10pt]
  \normalsize $^{1}$\textit{RUZIVO Research Lab, African Leadership University, Kigali, Rwanda}\\[2pt]
  \normalsize $^{2}$\textit{Stanford University, Stanford, CA, USA}\\[2pt]
  \normalsize $^{3}$\textit{Mila -- Quebec AI Institute and McGill University, Montreal, Canada}%
}

\date{}

\begin{document}
\maketitle

\begingroup
\renewcommand\thefootnote{}\footnotetext{\raggedright $^{*}$Corresponding author: \href{mailto:ocheankeli99@gmail.com}{ocheankeli99@gmail.com}\\
Code: \url{https://github.com/Tripp808/hausa-medical-asr}\\
Trained models: \url{https://huggingface.co/OcheAnkeli/hausa-medical-asr}}
\endgroup

\begin{abstract}
Automatic speech recognition (ASR) systems often struggle with specialised medical vocabulary, especially in low-resource languages where domain-specific speech data are scarce. We investigate text-only domain adaptation for Hausa medical ASR using the fixed MMS-1B-FL102 acoustic model and two decoding-time language-model fusion methods: Shallow Fusion (SF) and the Density Ratio Approach (DRA). A 4-gram KenLM trained on 19,421 Hausa health sentences serves as the medical target model, while a 75,372-sentence general-domain model provides the source prior for DRA. Evaluation uses 250 recordings of 25 physician-reviewed prompts from 10 speakers across six northern Nigerian states. Under prompt-disjoint five-fold cross-validation, SF and DRA reduce WER from 29.56\% to 23.55\% and 23.33\%, respectively. Both significantly outperform greedy decoding ($p<0.001$), while their difference is not significant ($p=0.423$). However, general-LM-only SF achieves 23.96\% WER, indicating that the medical LM's additional improvement over a general LM is modest. Medical Term Error Rate also improves under fusion, but does not establish a distinct medical-domain advantage. These findings demonstrate that lightweight, text-only language-model fusion can improve Hausa medical ASR without additional domain-specific speech collection.
\end{abstract}
\vspace{6pt} 


\section{Introduction}

Healthcare delivery across northern Nigeria -- home to a large Hausa-speaking population -- frequently relies on verbal communication for patient consultations, medication instructions, and clinical documentation. Automatic speech recognition capable of accurately transcribing medical Hausa speech would have direct and meaningful impact: supporting clinical documentation, telehealth, and patient education in resource-constrained healthcare settings where literacy in written Hausa is not universal.

Hausa is one of Africa's most widely spoken languages, with an estimated 94 million first- and second-language speakers across Nigeria, Niger, Ghana, Sudan, and other parts of West Africa \citep{ethnologue2025hausa}. Despite growing interest in Hausa natural language processing, reflected in the growing availability of Hausa language resources and benchmarks, including HausaNLP \citep{muhammad2025hausanlp}, Mozilla Common Voice \citep{ardila2020commonvoice}, and FLEURS \citep{conneau2023fleurs}, we are not aware of prior published work specifically addressing medical-domain ASR for Hausa.

The central bottleneck is audio data scarcity. Collecting paired audio-transcript data in specialised medical domains is expensive, logistically complex, and frequently infeasible. Fine-tuning acoustic models on domain-specific audio -- the standard approach for high-resource languages -- is not a viable option when such recordings do not exist at scale. This motivates a practical question: can domain adaptation for medical ASR be achieved using only text?

We answer affirmatively. We propose a text-only domain adaptation framework combining two language-model fusion strategies -- Shallow Fusion (SF) and the Density Ratio Approach (DRA) \citep{mcdermott2019density} -- with the MMS-1B-FL102 multilingual acoustic model \citep{pratap2024scaling}. We position our approach in contrast to the emerging use of large language models (LLMs) for ASR post-processing \citep{ma2023llmcorrection}. While LLM-based correction can improve transcription quality by exploiting broader linguistic context, text-only post-processing operates after acoustic decoding and therefore cannot directly recover information absent from the ASR output. Moreover, reliance on large generative models introduces substantially greater computational and deployment requirements than our lightweight decoding-time approach: our KenLM-based fusion relies on a compact 16.76~MB model.

We evaluate with sentence-prompt-level 5-fold cross-validation: language-model and decoding hyperparameters are selected only on each fold's tuning prompts and frozen before decoding that fold's held-out prompts, and the five held-out sets are pooled into 250 out-of-fold (OOF) predictions, so no evaluation prompt is used for tuning in the fold in which it is evaluated.

\noindent The contributions of this work are:
\begin{itemize}[leftmargin=1.2em,itemsep=1pt,topsep=2pt]
\item A text-only domain adaptation framework for Hausa medical ASR using SF and DRA, achieving 20.3\% and 21.1\% relative WER reductions, respectively, over an unadapted multilingual baseline, under sentence-level 5-fold cross-validation.
\item A medical-domain Hausa language model, consisting of a 4-gram KenLM trained on 19{,}421 sentences from the BBC Hausa and Deutsche Welle Hausa health sections, and a general-domain 75{,}372-sentence source language model enabling density-ratio correction of the acoustic model's language prior.
\item A purpose-built Hausa medical speech evaluation dataset of 250 utterances: 25 prompts independently reviewed by two medical doctors, each recorded by 10 speakers across six northern Nigerian states.
\item Medical Term Error Rate (\mter{}), a domain-specific metric for evaluating recognition of medical vocabulary, over a 66-term lexicon spanning six clinical categories, reported together with its construction caveats.
\item A domain contrast analysis and a cross-validated significance testing protocol (cluster bootstrap over sentence prompts) that together support the reported gains within the stated limits.
\end{itemize}

All trained language models and code are released publicly to support reproducibility and future benchmarking.

\section{Related Work}

\subsection{Domain Adaptation for ASR}
Domain adaptation in ASR broadly includes acoustic model adaptation, which typically relies on speech data, and language model adaptation, which can leverage text-only domain resources \citep{yadav2022survey} and is particularly suited to low-resource settings where domain-specific audio is unavailable.

For end-to-end ASR, Shallow Fusion combines external LM log-probabilities with ASR scores during beam-search decoding \citep{chorowski2017towards,gulcehre2017integrating} and has been shown to improve recognition of domain-specific vocabulary \citep{kannan2018analysis}. A known limitation is that the external LM may duplicate information already encoded by the acoustic model \citep{mcdermott2019density} -- particularly relevant for large multilingual models such as MMS, trained across more than 1{,}000 languages \citep{pratap2024scaling}.

The Density Ratio Approach (DRA) \citep{mcdermott2019density} addresses this using a Bayesian formulation: rather than simply adding a target-domain LM, DRA incorporates a source-domain LM to approximate the prior already encoded by the acoustic model, reducing double-counting. To our knowledge, no prior work has applied DRA to Nigerian-language ASR or medical ASR in an African language.

\subsection{LLM Post-Processing vs.\ Decoding-Time Fusion}
A growing body of work explores LLMs as ASR post-processors, correcting transcription errors by exploiting broader linguistic context \citep{ma2023llmcorrection}. This operates after acoustic decoding and cannot recover information absent from the ASR hypothesis -- particularly limiting in low-resource medical ASR, where terminology may be poorly represented in both the acoustic model and the decoded hypothesis. Deploying large generative models also introduces substantially greater computational and latency requirements than lightweight n-gram fusion. Our approach instead integrates the LM during beam-search decoding, using a compact 16.76~MB KenLM, substantially more suitable for resource-constrained deployment.

\subsection{ASR for African and Low-Resource Languages}
Multilingual self-supervised speech models such as XLS-R \citep{babu2022xlsr} and MMS \citep{pratap2024scaling} have substantially expanded ASR coverage for low-resource languages, including Hausa. Community efforts such as Masakhane \citep{nekoto2020participatory} and HausaNLP \citep{muhammad2025hausanlp} have further advanced datasets, benchmarks, and language technologies for African languages, and recent work has explored multilingual ASR for Kinyarwanda, Swahili, and Luganda \citep{elamin2023multilingual}. AfriSpeech-200 \citep{olatunji2023afrispeech} introduced Pan-African accented English speech spanning clinical and general domains, but focuses on English rather than African languages. For Hausa specifically, existing speech resources primarily target general-domain speech; to our knowledge, no prior published study has addressed medical-domain ASR for Hausa.

\subsection{Medical ASR and Terminology-Aware Evaluation}
Medical ASR has been extensively studied in high-resource settings, where specialised terminology poses challenges for general-domain ASR and motivates domain-specific LM adaptation \citep{afonja2024performant}. These studies, along with analyses of dictated clinical documents \citep{zhou2018analysis}, highlight the limitations of aggregate WER for clinically important terminology, motivating complementary domain-specific evaluation. We adapt this principle to Hausa with \mter{} (Section~\ref{sec:metrics}).

\section{Methodology}

\subsection{Acoustic Model}
We evaluated several multilingual ASR models on Hausa speech, including Whisper variants and Wav2Vec2-based models. MMS-1B-FL102 \citep{pratap2024scaling} was selected as our acoustic backbone for its consistently strong baseline (greedy WER 29.56\%). MMS provides a dedicated Hausa CTC adapter (language code \textit{hau}), enabling direct inference without fine-tuning; because CTC-based models lack an autoregressive language decoder, external LM integration during beam search is a natural fusion mechanism. MMS integrates directly with \textit{pyctcdecode} \citep{kensho_pyctcdecode} for SF and DRA decoding; audio is resampled to 16~kHz mono via \textit{ffmpeg}. The acoustic model is never fine-tuned, selected, or retuned on any fold: it is held fixed throughout, so reported gains are attributable to language-model adaptation alone.

\subsection{Medical Language Model (Target LM)}
We constructed a Hausa medical text corpus by scraping health-section articles from BBC Hausa (766 articles, 532{,}775 words) and Deutsche Welle Hausa (11 articles, 3{,}946 words), then applied Unicode normalisation, lowercasing, punctuation removal (preserving diacriticals), whitespace normalisation, deduplication, and length filtering (minimum 3 words), yielding 19{,}421 sentences ($\sim$558{,}700 words).

A 4-gram KenLM was trained via \textit{lmplz} \citep{heafield2013scalable} with modified Kneser--Ney smoothing \citep{chen1999empirical}, completing in 2.5s on CPU and producing a 16.76~MB binary. Held-out perplexity (1{,}943 sentences) was 101.86.

\subsection{General Language Model (Source LM)}
DRA requires a source LM approximating the acoustic model's implicit general-domain language prior \citep{mcdermott2019density}. We built a 75{,}372-sentence corpus from four public Hausa sources spanning formal and informal registers: Common Voice Hausa \citep{ardila2020commonvoice}; HausaVisualGenome \citep{abdulmumin2022hausavisualgenome} (32{,}923 post-edited image descriptions); NaijaSenti-Twitter Hausa \citep{muhammad2022naijasenti} ($\sim$14{,}000 annotated tweets); and the ijdutse corpus \citep{inuwadutse2021first} (general Hausa text, language-filtered). Held-out perplexity was 47.70.

\subsection{Domain Contrast Validation}
\label{sec:domaincontrast}
We assessed domain contrast by scoring every sentence in the evaluation set's 25 unique reference prompts against both the medical and general language models, and comparing these scores against 25 held-out general-domain sentences not used to train either LM. For each sentence, we computed the difference between the medical and general LM scores, $\log P_{\text{MED}} - \log P_{\text{GEN}}$. The medical references yielded a mean ratio of $+3.27$, while the held-out general sentences yielded a mean ratio of $-18.04$; a Mann--Whitney $U$ test confirmed this difference was statistically significant ($p<0.001$). This directional contrast supports the domain-specific probability ratio used by DRA. Illustrative examples are given in Appendix~\ref{app:additional}.

\subsection{Decoding Strategies}
Three decoding configurations are evaluated. \textbf{Greedy}: argmax CTC decoding with no language model, representing out-of-the-box MMS performance on Hausa medical speech. \textbf{Shallow Fusion (SF)}: beam search (width 100) combining CTC and medical KenLM log-probabilities,
\begin{equation}
\begin{aligned}
\log P_{\text{SF}}(W|X) = {}&\log P_{\text{CTC}}(W|X)\\
&+ \lambda \log P_{\text{MED}}(W) + \beta|W|.
\end{aligned}
\end{equation}
\textbf{Density Ratio (DRA)}: $n$-best rescoring ($n=50$) over the SF beam using
\begin{equation}
\begin{aligned}
\log P_{\text{DRA}}(W|X) = {}&\log P_{\text{CTC}}(W|X) + \lambda_\tau \log P_{\text{MED}}(W) \\
&- \lambda_\psi \log P_{\text{GEN}}(W) + \beta|W|.
\end{aligned}
\end{equation}
For each fold, SF hyperparameters are selected by grid search over $\lambda \in \{0.2,0.35,0.5,0.65,0.8,1.0\}$ and $\beta \in \{-1.0,-0.5,0.0,0.5,1.0,1.5\}$ on that fold's tuning prompts; DRA hyperparameters are selected over $\lambda_\tau \in \{0.3,0.5,0.8,1.0,1.2,1.5\}$, $\lambda_\psi \in \{0.1,0.3,0.5,0.8,1.0\}$ subject to $\lambda_\psi < \lambda_\tau$, and $\beta \in \{0.5,1.0,1.5\}$. Full per-fold selected values are given in Appendix~\ref{app:hyperparams}. Across folds, the selected SF weight is stable at $\lambda=0.8$, and DRA rescoring changes the SF hypothesis for 28 of the 250 (11.2\%) pooled out-of-fold utterances, consistent with DRA operating as a second-stage rescoring procedure over hypotheses already generated by SF.

\subsection{Cross-Validation Protocol}
\label{sec:cvprotocol}
We partition the 25 sentence prompts into five deterministic, sentence-disjoint folds. Each fold holds out 5 prompts (50 utterances, all 10 speakers) for evaluation and uses the remaining 20 prompts (200 utterances) for hyperparameter tuning. Language-model and decoding hyperparameters -- for SF, general-LM-only SF, and DRA alike -- are selected independently within each fold using only that fold's tuning prompts, then frozen before decoding that fold's held-out prompts. The five held-out sets are pooled to produce 250 out-of-fold (OOF) predictions, which are used for all reported metrics. No evaluation prompt is used for tuning in the fold in which it is evaluated. Because all ten speakers appear in both the tuning and held-out portions of every fold, this is prompt-level held-out evaluation, not a speaker-independent test.

Statistical uncertainty is estimated with a sentence-cluster bootstrap: 10{,}000 resamples are drawn at the level of sentence prompts (not individual utterances), with each resampled cluster retaining all ten speaker recordings for that prompt. This respects the nested structure of the data (25 sentences $\times$ 10 speakers) and yields more conservative confidence intervals than a flat utterance-level bootstrap (with only 25 prompt clusters, they remain indicative rather than high-power).

\section{Evaluation Dataset}

\subsection{Sentence Design}
We designed 25 evaluation sentences across three tiers of increasing medical vocabulary density: Tier 1 -- 8 short everyday health phrases (5--8 words); Tier 2 -- 12 medium clinical sentences (10--15 words); and Tier 3 -- 5 long, medically dense sentences (15--20 words). Initial sentence drafts were generated using an LLM (Claude Sonnet 5) and subsequently independently reviewed and corrected by two medical doctors for medical accuracy and Hausa naturalness. Full sentence lists are provided in Appendix~\ref{app:sentences}.

\subsection{Medical Expert Validation}
Two medical doctors -- Dr.\ Mohammed Mubarak Bello and Dr.\ Abdulmajid Imrana, both affiliated with Federal Neuro-Psychiatric Hospital, Kaduna, Nigeria -- independently reviewed all 25 sentences for medical accuracy and Hausa naturalness. Of the 25 sentences, 16 were approved verbatim and 9 received linguistic corrections concerning phrasing or register rather than medical accuracy. The corrected versions were used as the final recording references.

\subsection{Evaluation--Corpus Overlap}
To assess potential phrase-level overlap between the evaluation set and the medical LM training corpus, we measured 4-gram overlap across the 25 evaluation sentences. Nine sentences (36.0\%) contained at least one overlapping 4-gram, while 18 of 198 evaluation 4-grams (9.1\%) occurred in the training corpus. Thus, 90.9\% of evaluation 4-grams were not observed in the medical LM training corpus, limiting concerns that evaluation performance was driven primarily by direct phrase overlap.

\subsection{Speaker Recruitment and Recording}
\begin{figure}[!t]
\centering
\includegraphics[width=0.85\linewidth]{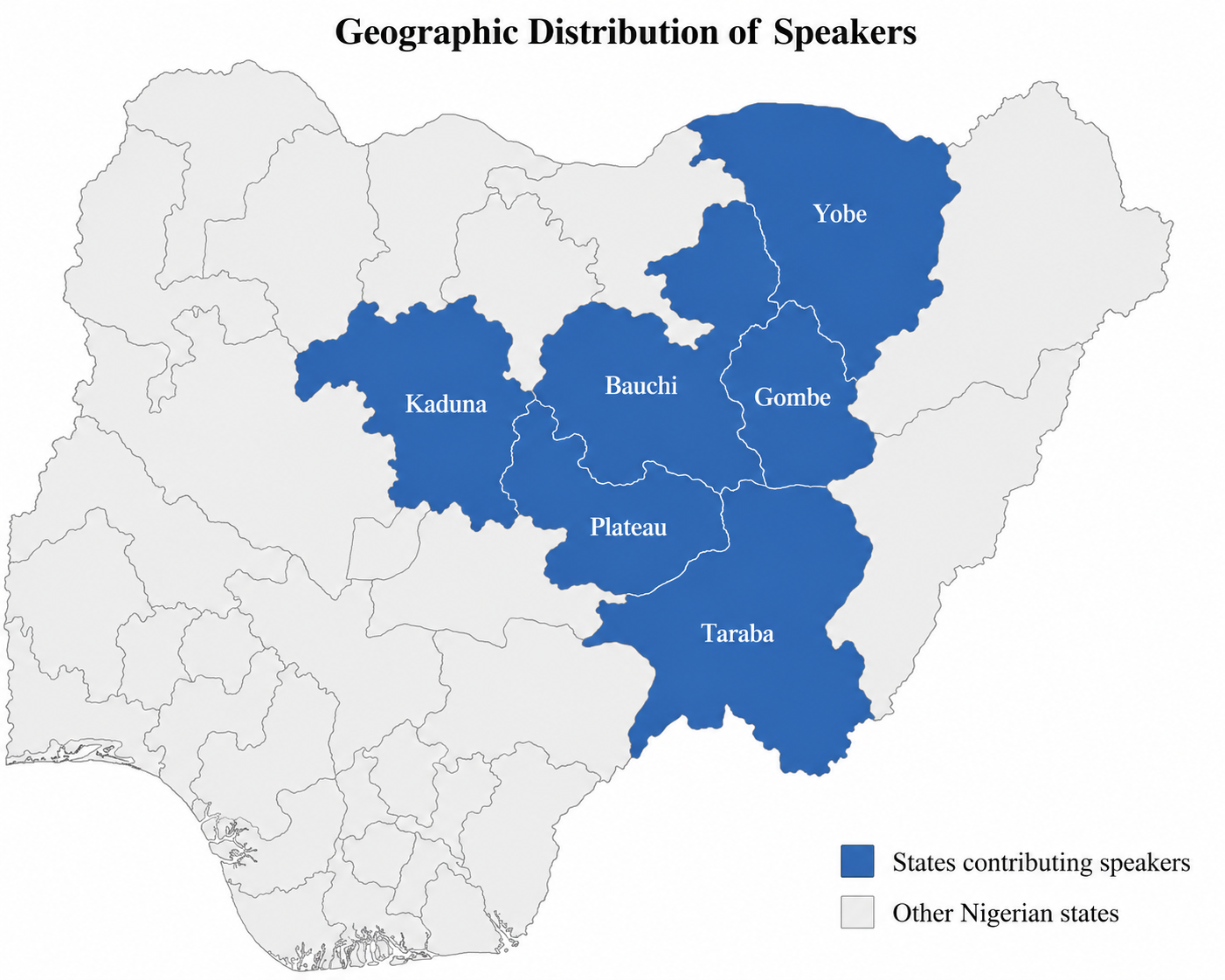}
\caption{Geographic distribution of speakers. States shaded contributed speakers to the evaluation corpus.}
\label{fig:speakermap}
\end{figure}

Ten volunteer speakers were recruited through community networks with emphasis on dialect diversity across northern Nigeria, representing six states: Plateau, Gombe, Taraba, Yobe, Kaduna, and Bauchi (Figure~\ref{fig:speakermap}). The cohort comprises 5 male and 5 female speakers, primarily aged 20--35, including students, NGO workers, and healthcare professionals. Approximately 8 of 10 are native Hausa speakers; 2 are fluent second-language speakers. Speaker demographics are given in Appendix~\ref{app:speakers}. Each speaker recorded all 25 sentences as separate audio files using personal mobile phones in quiet indoor environments. Files were converted to 16~kHz mono WAV using \textit{ffmpeg}. The final dataset comprises 250 utterances.

\section{Experiments and Results}

\subsection{Evaluation Metrics}
\label{sec:metrics}
We report: (1) Word Error Rate (WER); (2) Character Error Rate (CER); and (3) Medical Term Error Rate (\mter{}), a domain-specific metric measuring the proportion of medical vocabulary instances in the reference that are not correctly recognised in the hypothesis. \mter{} builds on prior domain-specific medical ASR metrics, including Medical Concept WER \citep{afonja2024performant}, which restrict error measurement to clinically relevant medical concepts or entities. Unlike these approaches, which rely on clinical NLP tools primarily available for English, \mter{} derives its medical lexicon from language-model statistics and expert validation, providing an approach applicable to low-resource languages without established clinical NLP tooling.

The lexicon was fixed before the cross-validated evaluation and lists 66 term strings (60 distinct after the diacritic folding ƙ/k, ɗ/d, ɓ/b used for matching) in six clinical categories (Symptoms \& Conditions, Medications \& Treatments, Clinical Procedures, Medical Personnel \& Settings, Anatomy, Clinical Descriptors). Of these, 48 are corpus-derived (medical-to-general LM unigram log-probability ratio $>0.5$), 58 are expert-validated, and 40 are both (Appendix~\ref{app:lexicon}). Because the corpus-derived criterion uses the same language models whose effect we measure, \mter{} is not fully independent of the medical LM; the 8 corpus-only terms account for 120 of the 790 reference instances (15.2\%).

Matching is occurrence-based and position-independent, after diacritic folding; morphological variants are distinct terms (no lemmatisation). \mter{} is computed as
\begin{equation}
\mter{} = \Big(1 - \frac{1}{N}\sum_{i=1}^{N} \mathbb{1}[t_i \in H]\Big) \times 100,
\end{equation}
where $t_i$ is a medical-term instance in the reference, $N$ is the total number of such instances (790, in 230 of the 250 utterances; prompts 006 and 020 contain no lexicon term), and $H$ is the set of (diacritic-folded) word forms in the hypothesis. \mter{} is recall-oriented and does not penalise inserted medical terms; it should be interpreted alongside WER and CER, never as a standalone measure of transcription quality.

\subsection{Main Results}
\label{sec:mainresults}
\begin{table}[!t]
\centering
\caption{Pooled out-of-fold performance over 250 utterances, 5-fold sentence-level cross-validation. 95\% confidence intervals for each system's WER are from the sentence-cluster bootstrap (10{,}000 resamples).}
\label{tab:main}
\footnotesize\setlength{\tabcolsep}{3pt}
\begin{adjustbox}{max width=\columnwidth}
\begin{tabular}{lrrrr}
\toprule
\textbf{System} & \textbf{WER (\%)} & \textbf{95\% CI} & \textbf{CER (\%)} & \textbf{Rel.\ WER $\downarrow$} \\
\midrule
Greedy (no LM) & 29.56 & [24.81, 35.24] & 6.86 & -- \\
Shallow Fusion & 23.55 & [19.26, 28.70] & 6.07 & 20.3\% \\
Density Ratio (DRA) & 23.33 & [19.14, 28.32] & 6.08 & 21.1\% \\
\bottomrule
\end{tabular}
\end{adjustbox}
\end{table}

\begin{figure}[!t]
\centering
\includegraphics[width=\linewidth]{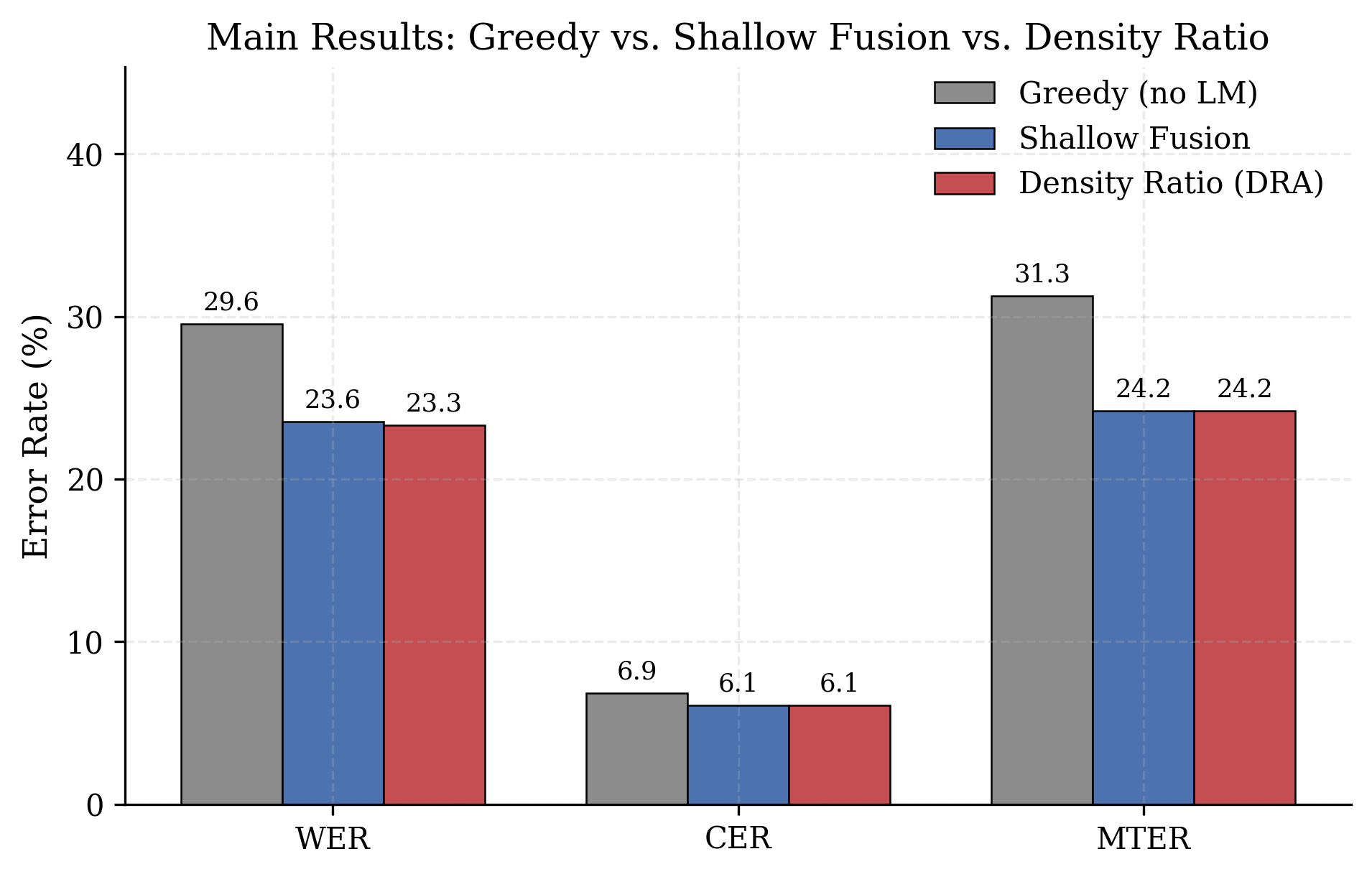}
\caption{Word Error Rate, Character Error Rate, and Medical Term Error Rate across Greedy, Shallow Fusion, and Density Ratio decoding, pooled out-of-fold.}
\label{fig:mainresults}
\end{figure}

Table~\ref{tab:main} reports pooled out-of-fold WER, CER, and relative improvement across the three decoding configurations (250 utterances). Both fusion methods significantly outperform greedy decoding: SF reduces WER by 6.01 points (20.3\% relative), with 95\% CI for the paired difference $[+4.04,+8.07]$ and $p<0.001$; DRA reduces WER by 6.23 points (21.1\% relative), CI $[+4.27,+8.31]$, $p<0.001$. DRA achieves the lowest point estimate, but its 0.22-point advantage over SF is not statistically significant (CI $[-0.25,+0.81]$, $p=0.423$). Thus, the evidence supports both fusion approaches substantially outperforming greedy decoding, not a statistically significant superiority of DRA over SF. Figure~\ref{fig:mainresults} visualises WER, CER, and \mter{} across all three configurations.

\subsection{Language-Model Ablation}
To assess whether the gains are specific to the medical domain, we also evaluated SF with the general-domain LM alone and with a combined general+medical LM, each with hyperparameters independently fold-tuned under the same protocol (Table~\ref{tab:ablation}). General-LM-only SF reaches 23.96\% WER -- only 0.41 points behind the medical-only LM (23.55\%) -- and the combined LM gives 23.85\%, no improvement over medical-only SF. On \mter{} the ordering reverses: general-only (23.80\%) and combined (23.67\%) are slightly lower than medical-only (24.18\%). Thus most of the improvement over greedy decoding comes from adding an external LM at all; the medical LM's additional benefit is small and shows up in WER rather than \mter{}.

\begin{table}[!t]
\centering
\caption{Shallow Fusion language-model ablation, fold-tuned under 5-fold CV (pooled out-of-fold).}
\label{tab:ablation}
\footnotesize\setlength{\tabcolsep}{3pt}
\begin{adjustbox}{max width=\columnwidth}
\begin{tabular}{lrr}
\toprule
\textbf{Configuration} & \textbf{WER (\%)} & \textbf{MTER (\%)} \\
\midrule
Greedy (no LM) & 29.56 & 31.27 \\
SF, general LM only & 23.96 & 23.80 \\
SF, medical LM only & 23.55 & 24.18 \\
SF, combined general + medical LM & 23.85 & 23.67 \\
\bottomrule
\end{tabular}
\end{adjustbox}
\end{table}

\subsection{Results by Sentence Tier}
\begin{figure}[!t]
\centering
\includegraphics[width=\linewidth]{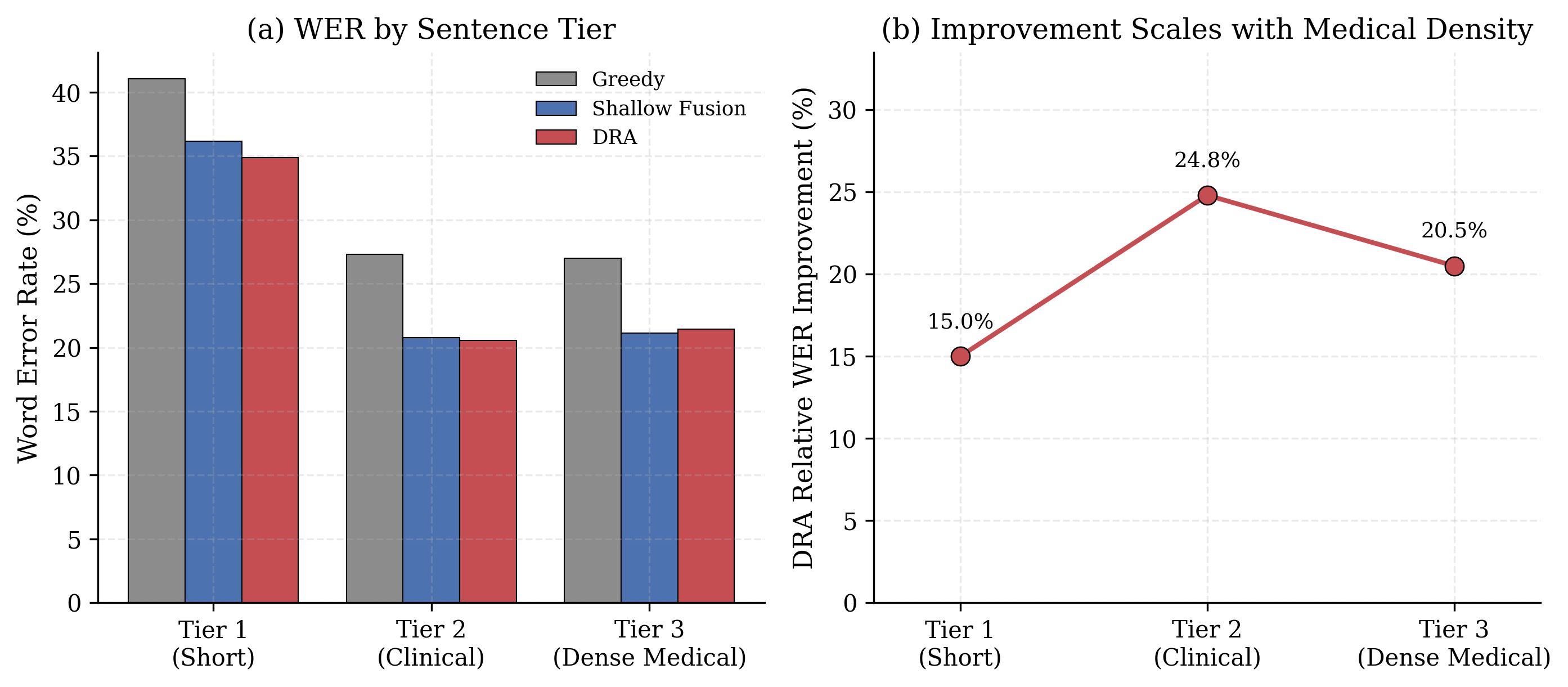}
\caption{(a) WER by sentence tier. (b) DRA's relative WER improvement over greedy, by tier.}
\label{fig:tier}
\end{figure}

Table~\ref{tab:tier} and Figure~\ref{fig:tier} report micro-averaged WER by sentence tier. All fusion methods outperform greedy decoding across all tiers. DRA's relative improvement over greedy is smallest for Tier~1 (short, low medical density; $+15.0\%$) and largest for Tier~2 (medium clinical density; $+24.8\%$), with Tier~3 (long, dense medical sentences) in between ($+20.5\%$). These differences may partly reflect sentence-length effects in addition to vocabulary density, so we do not attribute the tier gradient solely to medical vocabulary content.

\begin{table}[!t]
\centering
\caption{Micro-averaged WER by sentence tier, pooled out-of-fold.}
\label{tab:tier}
\footnotesize\setlength{\tabcolsep}{3pt}
\begin{adjustbox}{max width=\columnwidth}
\begin{tabular}{lrrrr}
\toprule
\textbf{Tier} & \textbf{N} & \textbf{Greedy} & \textbf{SF} & \textbf{DRA} \\
\midrule
Tier 1 (Short) & 80 & 41.06\% & 36.17\% & 34.89\% \\
Tier 2 (Clinical) & 120 & 27.31\% & 20.77\% & 20.54\% \\
Tier 3 (Dense Medical) & 50 & 26.98\% & 21.15\% & 21.46\% \\
\bottomrule
\end{tabular}
\end{adjustbox}
\end{table}

\subsection{MTER by Clinical Category}
\begin{table}[!t]
\centering
\caption{MTER by clinical category, pooled out-of-fold (790 term instances; category counts not mutually exclusive, sum to 800).}
\label{tab:mter_category}
\footnotesize\setlength{\tabcolsep}{3pt}
\begin{adjustbox}{max width=\columnwidth}
\begin{tabular}{lrrrr}
\toprule
\textbf{Category} & \textbf{N} & \textbf{Greedy} & \textbf{SF} & \textbf{DRA} \\
\midrule
Symptoms \& Conditions & 260 & 19.62\% & 14.62\% & 14.23\% \\
Medications \& Treatments & 150 & 48.00\% & 30.67\% & 30.67\% \\
Clinical Procedures & 70 & 24.29\% & 14.29\% & 14.29\% \\
Med.\ Personnel \& Settings & 60 & 6.67\% & 1.67\% & 1.67\% \\
Anatomy & 120 & 56.67\% & 57.50\% & 59.17\% \\
Clinical Descriptors & 140 & 25.71\% & 20.00\% & 19.29\% \\
\bottomrule
\end{tabular}
\end{adjustbox}
\end{table}

\begin{figure}[!t]
\centering
\includegraphics[width=\linewidth]{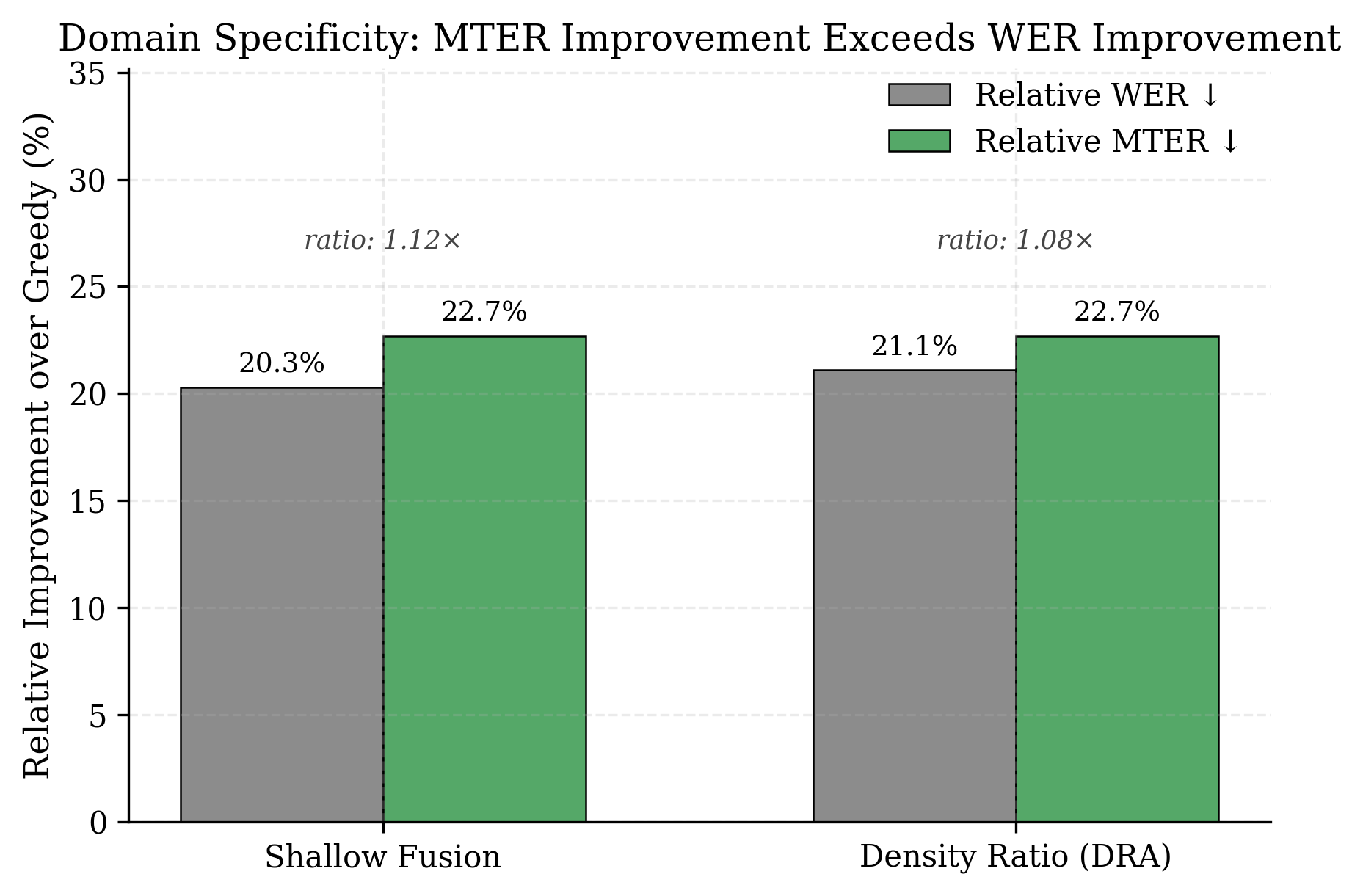}
\caption{Relative improvement in MTER consistently exceeds relative improvement in WER for both fusion methods.}
\label{fig:domainspec}
\end{figure}

Table~\ref{tab:mter_category} reports \mter{} across the six clinically motivated categories (790 medical-term instances in 230 of the 250 utterances; category-level counts sum to 800 because one term, \emph{tsanani}, belongs to two categories). DRA and SF reduce \mter{} most substantially in Medical Personnel \& Settings (from 6.67\% to 1.67\%) and Medications \& Treatments (from 48.00\% to 30.67\%), while Anatomy is the only category that does \emph{not} improve, and in fact worsens slightly under both fusion methods (56.67\% $\rightarrow$ 57.50\% SF, 59.17\% DRA), indicating that anatomy-related terminology remains challenging in the current adaptation setup, plausibly due to limited corpus representation and phonological overlap with everyday Hausa vocabulary. As Figure~\ref{fig:domainspec} shows, the relative \mter{} reduction (22.7\% for both methods) exceeds the relative WER reduction (20.3\% SF, 21.1\% DRA; ratios 1.12$\times$ and 1.08$\times$), but general-LM-only SF shows the same pattern (23.9\% vs.\ 18.9\% relative), so this gap does not by itself isolate a medical-domain effect (Section~\ref{sec:discussion}).

\subsection{Speaker-Level and Sex Analysis}
\begin{figure}[!t]
\centering
\includegraphics[width=\linewidth]{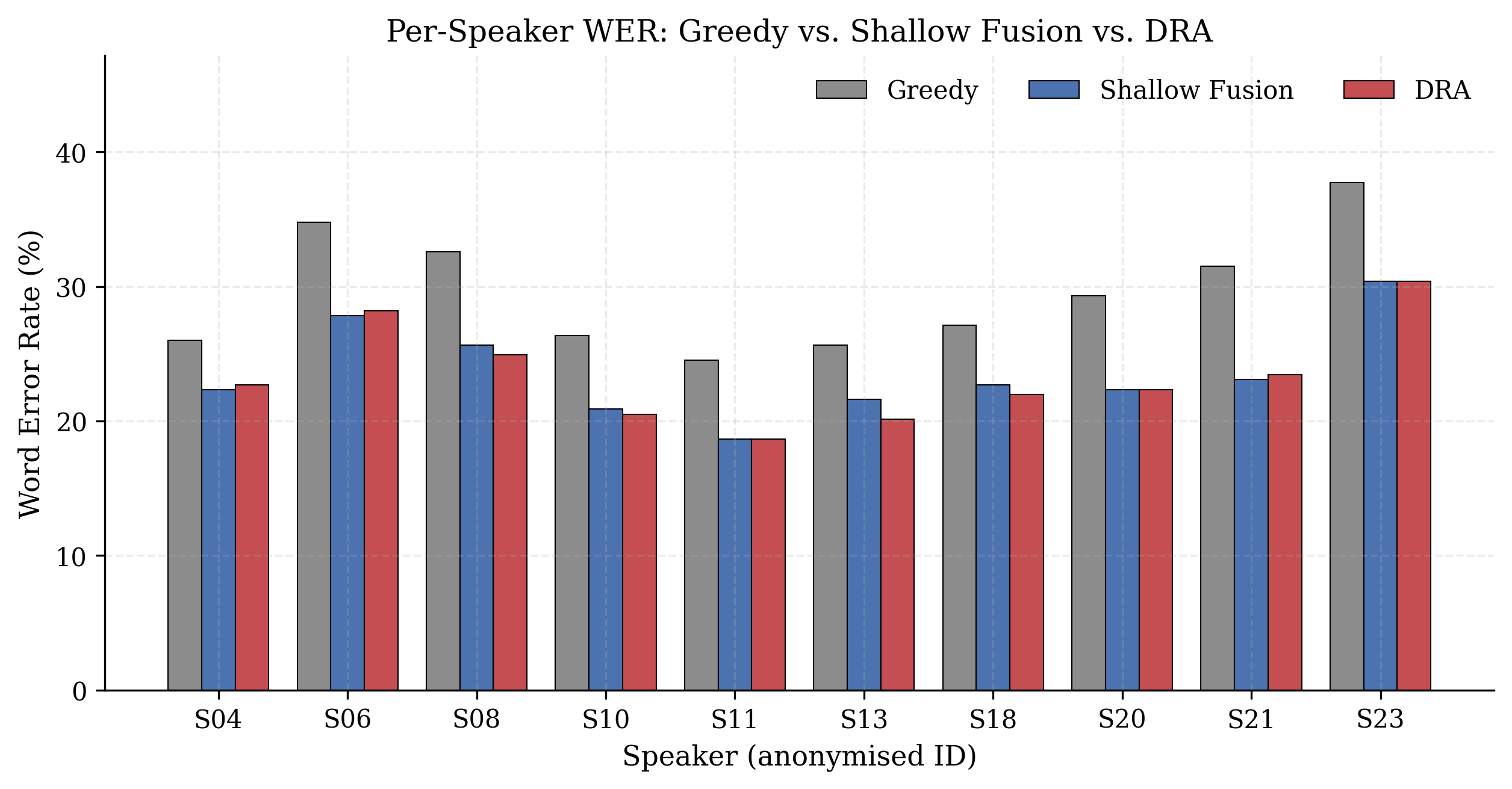}
\caption{Per-speaker Word Error Rate across all three decoding strategies, showing consistent improvement across the cohort.}
\label{fig:speakerwer}
\end{figure}

Figure~\ref{fig:speakerwer} shows per-speaker WER across all three decoding strategies (full values in Appendix~\ref{app:additional}, Table~\ref{tab:speakerwer_app}). All 10 speakers improve under at least one fusion method; three speakers (S04, S06, S21) show marginal DRA degradation relative to SF (at most 0.4 points). Dialect was not formally annotated in this study, so we do not attribute this to a specific phonological cause. Performance is similar across speaker sex: pooled WER is 28.86\%/23.08\%/23.00\% (Greedy/SF/DRA) for male speakers and 30.26\%/24.03\%/23.66\% for female speakers -- a difference of at most 1.40 points at any stage, and 0.66 points for DRA specifically. We do not perform a formal significance test on this comparison given the small number of speakers per group.

\section{Discussion}
\label{sec:discussion}

\subsection{On Text-Only Adaptation Effectiveness}
The primary finding is that text-only domain adaptation is effective and accessible for Hausa medical ASR, and that this holds under a properly cross-validated protocol. The pipeline -- from corpus collection to a deployment-ready LM binary -- requires no GPU, no proprietary data, and trains in seconds, offering a reproducible path to domain-adapted ASR for other low-resource languages with available domain text.

\subsection{MTER and the Role of the General LM}
\mter{} falls more than WER (Section~\ref{sec:mainresults}), but this does not by itself indicate domain-specific adaptation: general-LM-only SF attains a comparable, nominally lower \mter{} (23.80\% vs.\ 24.18\% for the medical LM), and the corpus-derived part of the lexicon is itself selected with the medical/general LM ratio. The defensible reading is that external-LM fusion improves terminology recall generally, with the medical LM's own contribution -- 0.41 WER points -- small and not isolated by \mter{}; an independently constructed or precision-aware lexicon would test this more directly.

\subsection{DRA vs.\ Shallow Fusion}
We treat the SF--DRA comparison as secondary: DRA has the lowest point WER, but its advantage is not significant under the cluster bootstrap, and DRA only rescores SF's own beam (changing 28 of 250, 11.2\%, of hypotheses). The 5-fold, prompt-disjoint protocol (Section~\ref{sec:cvprotocol}) makes every reported prediction out-of-fold with respect to prompt-level tuning and estimates uncertainty at the sentence-prompt level; with only 25 prompt clusters and 10 speakers, however, the intervals should be read as indicative. At 23.33\% WER, the system suits human-reviewed documentation workflows, not unsupervised clinical use.

\section{Limitations}

We discuss limitations across several dimensions. \textbf{Evaluation scale and speech style}: 250 utterances from 10 speakers is small, and the prompts are purpose-built sentences read aloud rather than spontaneous clinical conversation; results characterise this constructed benchmark, and broader generalisation requires larger, more diverse, naturalistic data. \textbf{Geographic coverage}: speakers are drawn from six northern Nigerian states; North-West varieties (Kano, Sokoto, Zamfara) and Niger Republic Hausa are underrepresented. \textbf{Speaker demographics}: speakers are predominantly young adults (20s--30s); performance on elderly speakers, children, and varying Hausa education levels is unknown. \textbf{Fold independence}: only sentences, not speakers, are held out within each fold -- all ten speakers appear in both tuning and held-out portions of every fold; a speaker-disjoint CV would be a complementary, stronger generalisation test. \textbf{N-gram LM limitations}: 4-gram KenLMs capture local statistics but not long-range coherence; neural LMs would likely help further, particularly for Tier~3. \textbf{Source LM approximation}: our 75{,}372-sentence source LM is far smaller than the acoustic model's own pretraining corpus, likely capping how well it approximates MMS's true prior and how large a DRA advantage over SF is achievable. \textbf{\mter{} scope}: \mter{} uses diacritic-folded, position-independent term matching over a lexicon partly derived from the LM probability ratio, does not model morphological equivalence, and does not penalise inserted terms; it complements, rather than replaces, a full clinical-correctness measure. \textbf{Data ethics}: BBC/DW Hausa text was scraped for non-commercial research; only the trained LM binary, not raw text, will be released. Similarly, due to speaker privacy, the evaluation audio recordings are not publicly released.

\section{Ethical Statement}

The study adopted a division of labour in which model development and evaluation were author-led, while medical validation of the evaluation sentences was conducted independently by two Hausa-speaking medical doctors. The clinicians reviewed the sentences for medical accuracy, Hausa naturalness, and recording clarity, with their corrections incorporated before recording. This process provided domain and linguistic validation for a study whose modelling was conducted by researchers who are not first-language Hausa speakers. All volunteer speakers provided informed consent and were informed of the research purpose and intended dataset release. Speakers are identified by anonymised IDs in all published materials. Only the the trained language models and the code are made available. To protect speaker privacy, the raw audio recordings and evaluation dataset are not publicly released. The resulting models and resources are released openly to support further research and downstream applications; they should be regarded as research resources rather than clinically validated systems, and deployment in healthcare settings would require additional clinical, demographic, and regulatory validation. The system makes no claim of clinical validation and is not suitable for unsupervised clinical decision-making, including unreviewed medication or dosage transcription.

% \paragraph{AI Use Disclosure.} AI tools were used to assist with editing and improving the clarity of the manuscript text and to assist with figure creation. All AI-assisted content was reviewed and verified by the authors.

\section*{Acknowledgments}

The authors thank Dr.\ Mohammed Mubarak Bello and Dr.\ Abdulmajid Imrana (Federal Neuro-Psychiatric Hospital, Kaduna, Kaduna State, Nigeria) for independent medical expert validation, and the ten volunteer speakers for their recordings and patience. We further thank the following contributors for data collection, annotation, and project coordination: Ismail Kewa, Chukwuma Onuoha, Yakubu Emmanuel Yakubu, Ahmad Daib, Ishaq Adamu, Patience Habila Mugana, Esther Isah Bila, Hafsat Abubakar Ahmad, Fatima Mudi Abdullahi, and Fatima Ahmed. This research was conducted within the RUZIVO Research Lab at African Leadership University. The authors declare no competing interests and received no dedicated external funding for this work.

\section{Conclusion}

We have presented a text-only domain adaptation framework for Hausa medical ASR, achieving 20.3\% (SF) and 21.1\% (DRA) relative WER reductions over an unadapted MMS baseline under sentence-prompt-level 5-fold cross-validation, with no evaluation prompt used for tuning in its own fold. Most of the gain comes from adding an external LM: general-LM-only SF already reaches 23.96\% WER, the medical LM adds a smaller 0.41-point benefit, and DRA's advantage over SF is not statistically significant. \mter{} falls 22.7\% under both methods, but general-LM-only fusion gives a comparable \mter{}, so we treat it as complementary evidence. The pipeline produces a compact 16.76~MB model, practical for resource-constrained deployment. We contribute a Hausa medical speech evaluation corpus built from 25 physician-reviewed prompts and introduce \mter{}. We release all trained models and code to support future Hausa and African-language medical ASR work.

\bibliographystyle{unsrtnat}
\bibliography{references}

\onecolumn
\appendix

\section{Speaker Demographics}
\label{app:speakers}

Table~\ref{tab:speakerdemo} presents the anonymised demographic summary of the 10 volunteer speakers. Speaker IDs are used throughout all published materials and released datasets. No personally identifying information is associated with speaker IDs in any public release.

\begin{table}[H]
\centering
\caption{Anonymised speaker demographics. 8 of 10 speakers are native Hausa speakers; 2 are fluent second-language speakers.}
\label{tab:speakerdemo}
\small
\begin{tabular}{llllll}
\toprule
\textbf{Speaker ID} & \textbf{State} & \textbf{Region} & \textbf{Sex} & \textbf{Hausa Proficiency} & \textbf{Occupation} \\
\midrule
S04 & Plateau & North-Central & M & Native & NGO / Community \\
S06 & Gombe   & North-East    & M & Fluent (L2) & Student \\
S08 & Taraba  & North-East    & M & Native & Student \\
S10 & Yobe    & North-East    & M & Native & NGO / Community \\
S11 & Kaduna  & North-West    & M & Native & Student \\
S13 & Gombe   & North-East    & F & Native & Student \\
S18 & Gombe   & North-East    & F & Native & Healthcare Worker \\
S20 & Bauchi  & North-East    & F & Native & Student \\
S21 & Bauchi  & North-East    & F & Native & NGO / Community \\
S23 & Gombe   & North-East    & F & Fluent (L2) & Healthcare Worker \\
\bottomrule
\end{tabular}
\end{table}

\section{Evaluation Sentences}
\label{app:sentences}

All 25 evaluation sentences are listed below, grouped by tier. Sentences marked $\ast$ received linguistic corrections from medical expert annotators; all others were approved verbatim. All sentences are presented in their final validated form as used for recordings.

\begin{longtable}{p{0.8cm}p{0.8cm}p{6.5cm}p{6.5cm}p{0.5cm}}
\caption{All 25 evaluation sentences in final validated form. $\ast$ = linguistically corrected by medical annotators.}\label{tab:evalsentences}\\
\toprule
\textbf{ID} & \textbf{Tier} & \textbf{Hausa Sentence (Validated)} & \textbf{English Translation} & \textbf{$\ast$} \\
\midrule
\endfirsthead
\toprule
\textbf{ID} & \textbf{Tier} & \textbf{Hausa Sentence (Validated)} & \textbf{English Translation} & \textbf{$\ast$} \\
\midrule
\endhead
\midrule
\multicolumn{5}{r}{\footnotesize Continued on next page}\\
\endfoot
\bottomrule
\endlastfoot
001 & 1 & Ka sha ruwa mai yawa kowace rana. & Drink plenty of water every day. & \\
002 & 1 & Ina jin zafi a kirjina. & I feel pain in my chest. & \\
003 & 1 & Jikinna ya kama zafi yau sosai. & My body is very hot today. & $\ast$ \\
004 & 1 & Ina buƙatar magani don ciwon kai. & I need medicine for a headache. & \\
005 & 1 & Ka huta sosai don ka warke da wuri. & Rest well for a quick recovery. & $\ast$ \\
006 & 1 & Ina jin kishin ruwa sosai. & I am feeling very thirsty. & $\ast$ \\
007 & 1 & Hannuwana na karkarwa. & My hands are shaking. & $\ast$ \\
008 & 1 & Ina tai amai sau da yawa yau. & I have been vomiting many times today. & $\ast$ \\
009 & 2 & Likita ya ce ina da zazzabin cizon sauro kuma ya rubuta magani. & The doctor said I have malaria and prescribed medicine. & \\
010 & 2 & Ka sha ƙwayar maganin sau uku a kowace rana bayan abinci. & Take the tablet three times daily after meals. & \\
011 & 2 & An diba jinina a asibiti domin gwaji. & Blood was taken from me at the hospital for testing. & $\ast$ \\
012 & 2 & Ciwon sukari yana buƙatar kulawa ta musamman da abinci mai kyau. & Diabetes requires special care and a good diet. & \\
013 & 2 & Ina da tari mai tsanani tun sati ɗaya da ya wuce. & I have had a severe cough for the past one week. & \\
014 & 2 & Likita ya umurce ni da in dawo bayan kwana bakwai don duba. & The doctor told me to return after seven days for a checkup. & \\
015 & 2 & Yaron yana da gudawa kuma jikansa ya yi rauni sosai. & The child has diarrhea and his body has become very weak. & \\
016 & 2 & An gaya mini cewa matsin jinina ya yi yawa fiye da yadda ya kamata. & I was told that my blood pressure is higher than it should be. & \\
017 & 2 & Ina bukatar allurar rigakafi kafin in tafi in nayi tafiya kasar waje. & I need a vaccination before traveling abroad. & $\ast$ \\
018 & 2 & Tana jin zafi a ƙashin baya kuma ba ta iya tsayawa tsawon lokaci. & She feels pain in her lower back and cannot stand for long. & \\
019 & 2 & Likita ya rubuta mun magunguna guda uku daban-daban. & The doctor gave three different written prescriptions. & $\ast$ \\
020 & 2 & Muna roƙonka ka wanke hannuwanka kafin ka taɓa yaro. & We ask you to wash your hands before touching the child. & \\
021 & 3 & Bayan gwajin jini, an gano cewa matakan sukari a jikina sun fi iyaka kuma ana buƙatar magani na musamman. & After the blood test, sugar levels were found to exceed the limit and special medication is needed. & \\
022 & 3 & Alamun cutar zazzabin cizon sauro sun haɗa da zazzabi mai ƙarfi, ciwon kai, rawar jiki, da rashin ƙarfi gabaɗaya. & Symptoms of malaria include high fever, headache, body shaking, and general weakness. & \\
023 & 3 & An shawarce ni da in guji abinci mai yawan kise da gishiri saboda matsalar zuciyata. & I was advised to avoid foods high in fat and salt because of my heart condition. & $\ast$ \\
024 & 3 & Dole ne a kai yaron asibiti da sauri saboda yana da alamu na rashin ruwa a jiki wanda ke da haɗari. & The child must be taken to the hospital quickly because he shows signs of dehydration which is dangerous. & \\
025 & 3 & Likita ya bayyana cewa zan buƙaci yin tiyata idan magunguna na yau da kullum ba su yi aiki ba cikin makonni biyu. & The doctor explained that surgery will be needed if regular medications do not work within two weeks. & \\
\end{longtable}

\section{Medical Term Lexicon}
\label{app:lexicon}

The fixed lexicon used for \mter{} lists 66 term strings (60 distinct after diacritic folding), grouped by clinical category. Corpus-derived terms ($\dagger$; 48) were identified by medical-to-general LM unigram log-probability ratio $>0.5$; because this criterion is computed from the two LMs being compared, these terms are not an independent signal of medical-LM quality (Section~\ref{sec:metrics}). Expert-validated terms ($\ddagger$; 58) were confirmed by medical annotators as clinically significant regardless of corpus frequency ratio; 40 terms are both. The 8 corpus-only terms are \emph{bukaci, bukatar, buƙaci, buƙatar, kise, likita, rauni, zuciyata}. Spelling variants with and without hooked letters are listed separately but match identically after folding.

\begin{longtable}{p{3.8cm}p{12.7cm}}
\caption{Medical term lexicon: 66 term strings across six clinical categories. \emph{tsanani} appears in two categories (Symptoms \& Conditions and Clinical Descriptors), so category slots total 66 while only 65 terms are categorised; \emph{lafiya} is expert-validated but not assigned to a category (and does not occur in the 25 reference sentences).}\label{tab:lexicon}\\
\toprule
\textbf{Category (terms)} & \textbf{Terms ($\dagger$=corpus-derived, $\ddagger$=expert-validated)} \\
\midrule
\endfirsthead
\toprule
\textbf{Category (terms)} & \textbf{Terms ($\dagger$=corpus-derived, $\ddagger$=expert-validated)} \\
\midrule
\endhead
\bottomrule
\endfoot
Symptoms \& Conditions (22) & amai$^{\dagger\ddagger}$ ciwo$^{\dagger\ddagger}$ ciwon$^{\dagger\ddagger}$ cizon$^{\dagger\ddagger}$ cutar$^{\dagger\ddagger}$ gudawa$^{\dagger\ddagger}$ hadari$^{\dagger\ddagger}$ haɗari$^{\dagger\ddagger}$ karkarwa$^{\dagger\ddagger}$ rashin$^{\dagger\ddagger}$ rawar$^{\dagger\ddagger}$ sauro$^{\dagger\ddagger}$ sukari$^{\dagger\ddagger}$ tari$^{\dagger\ddagger}$ tsanani$^{\dagger\ddagger}$ zazzabi$^{\dagger\ddagger}$ zazzabin$^{\dagger\ddagger}$ rauni$^{\dagger}$ hawan$^{\ddagger}$ kishirwa$^{\ddagger}$ zafi$^{\ddagger}$ ƙishirwa$^{\ddagger}$ \\
Medications \& Treatments (12) & allurar$^{\dagger\ddagger}$ kulawa$^{\dagger\ddagger}$ kwayar$^{\dagger\ddagger}$ magani$^{\dagger\ddagger}$ maganin$^{\dagger\ddagger}$ magunguna$^{\dagger\ddagger}$ rigakafi$^{\dagger\ddagger}$ tiyata$^{\dagger\ddagger}$ ƙwayar$^{\dagger\ddagger}$ kise$^{\dagger}$ sha$^{\ddagger}$ warke$^{\ddagger}$ \\
Clinical Procedures (6) & gwaji$^{\dagger\ddagger}$ gwajin$^{\dagger\ddagger}$ dawo$^{\ddagger}$ diba$^{\ddagger}$ duba$^{\ddagger}$ rubuta$^{\ddagger}$ \\
Medical Personnel \& Settings (2) & asibiti$^{\dagger\ddagger}$ likita$^{\dagger}$ \\
Anatomy (12) & jikinna$^{\dagger\ddagger}$ jini$^{\dagger\ddagger}$ jinina$^{\dagger\ddagger}$ kashin$^{\dagger\ddagger}$ ƙashin$^{\dagger\ddagger}$ baya$^{\ddagger}$ hannuwana$^{\ddagger}$ jikansa$^{\ddagger}$ jiki$^{\ddagger}$ jikina$^{\ddagger}$ kirji$^{\ddagger}$ kirjina$^{\ddagger}$ \\
Clinical Descriptors (12) & alamu$^{\dagger\ddagger}$ iyaka$^{\dagger\ddagger}$ makonni$^{\dagger\ddagger}$ matakan$^{\dagger\ddagger}$ matsin$^{\dagger\ddagger}$ musamman$^{\dagger\ddagger}$ tsanani$^{\dagger\ddagger}$ bukaci$^{\dagger}$ bukatar$^{\dagger}$ buƙaci$^{\dagger}$ buƙatar$^{\dagger}$ zuciyata$^{\dagger}$ \\
Not assigned to a category (1) & lafiya$^{\ddagger}$ \\
\end{longtable}

\section{Hyperparameter Selection}
\label{app:hyperparams}

\subsection{Grid Search Ranges}
SF hyperparameters were selected per fold by grid search over $\lambda \in \{0.2,0.35,0.5,0.65,0.8,1.0\}$ and $\beta \in \{-1.0,-0.5,0.0,0.5,1.0,1.5\}$. The general-LM-only SF ablation used the same grid over the general-domain LM. DRA hyperparameters were selected per fold over $\lambda_\tau \in \{0.3,0.5,0.8,1.0,1.2,1.5\}$, $\lambda_\psi \in \{0.1,0.3,0.5,0.8,1.0\}$ subject to $\lambda_\psi < \lambda_\tau$, and $\beta \in \{0.5,1.0,1.5\}$, evaluated via $n$-best rescoring with $n=50$ over the fold-specific SF beam.

\subsection{Per-Fold Selected Values}
Table~\ref{tab:perfold} gives the selected hyperparameters for each of the five cross-validation folds, each tuned only on that fold's 20 tuning prompts (200 utterances) and frozen before decoding that fold's 5 held-out prompts (50 utterances).

\begin{table}[H]
\centering
\caption{Per-fold selected hyperparameters. SF~$\lambda$/$\beta$: medical-LM Shallow Fusion weight/word score. Gen.~$\lambda$/$\beta$: general-LM-only SF ablation. DRA $\lambda_\tau$/$\lambda_\psi$/$\beta$: Density Ratio weights. DRA $n$-best = 50 throughout.}
\label{tab:perfold}
\small
\begin{tabular}{cccccccc}
\toprule
\textbf{Fold} & \textbf{Held-out prompts} & \textbf{SF $\lambda$} & \textbf{SF $\beta$} & \textbf{Gen.\ $\lambda$} & \textbf{Gen.\ $\beta$} & \textbf{DRA $\lambda_\tau$} & \textbf{DRA $\lambda_\psi$ / $\beta$} \\
\midrule
1 & 010, 016, 017, 020, 021 & 0.8 & 1.5 & 0.8 & 1.5 & 1.0 & 0.1 / 0.5 \\
2 & 007, 008, 011, 018, 025 & 0.8 & 0.0 & 0.8 & 1.5 & 1.2 & 0.1 / 0.5 \\
3 & 001, 004, 013, 019, 022 & 0.8 & 1.5 & 0.8 & 1.5 & 1.2 & 0.1 / 0.5 \\
4 & 003, 006, 012, 015, 024 & 0.8 & 1.5 & 0.8 & 1.5 & 1.2 & 0.1 / 0.5 \\
5 & 002, 005, 009, 014, 023 & 0.8 & 1.5 & 0.8 & 1.5 & 1.2 & 0.1 / 0.5 \\
\bottomrule
\end{tabular}
\end{table}

The SF weight $\lambda=0.8$ is selected consistently across all five folds, and DRA's $\lambda_\psi=0.1$ and $\beta=0.5$ are likewise stable, indicating that the optimal fusion strength is not an artefact of a particular tuning split. DRA's $\lambda_\tau$ varies slightly (1.0 in Fold~1, 1.2 elsewhere).

\FloatBarrier
\section{Additional Results}
\label{app:additional}

The domain-contrast analysis (Section~\ref{sec:domaincontrast}) is reported as aggregate statistics: mean $\log P_{\text{MED}}-\log P_{\text{GEN}}$ of $+3.27$ for the 25 medical prompts and $-18.04$ for 25 held-out general sentences (Mann--Whitney $U$, $p=4.1\times10^{-9}$).

Table~\ref{tab:speakerwer_app} gives the full per-speaker WER values underlying Figure~\ref{fig:speakerwer}.

\begin{table}[H]
\centering
\caption{Per-speaker pooled out-of-fold WER across all three decoding strategies.}
\label{tab:speakerwer_app}
\small
\begin{tabular}{lrrr}
\toprule
\textbf{Speaker} & \textbf{Greedy} & \textbf{SF} & \textbf{DRA} \\
\midrule
S04 & 26.01\% & 22.34\% & 22.71\% \\
S06 & 34.80\% & 27.84\% & 28.21\% \\
S08 & 32.60\% & 25.64\% & 24.91\% \\
S10 & 26.37\% & 20.88\% & 20.51\% \\
S11 & 24.54\% & 18.68\% & 18.68\% \\
S13 & 25.64\% & 21.61\% & 20.15\% \\
S18 & 27.11\% & 22.71\% & 21.98\% \\
S20 & 29.30\% & 22.34\% & 22.34\% \\
S21 & 31.50\% & 23.08\% & 23.44\% \\
S23 & 37.73\% & 30.40\% & 30.40\% \\
\bottomrule
\end{tabular}
\end{table}

\end{document}